\documentclass[runningheads]{llncs}
\usepackage{graphicx,algpseudocode,subfigure,wrapfig}
\usepackage[ruled]{algorithm2e}

\begin{document}

\title{Automated Neuron Shape Analysis from Electron Microscopy  \thanks{Supported by the Intelligence Advanced Research Projects Activity (IARPA) via Department of Interior/Interior Business Center (Dol/IBC) contract number D16PC00004. The U.S. Government is authorized to reproduce and distribute reprints for Governmental purposes nonwithstanding any copyright annotation theron. Disclamer: The views and conclusion contained herin are those of the authors and should not be interpreted as necessarily representing the official policies or endorsements, either expressed or implied , of IARPA, DoI/IBC, or the U.S. Government. We thank Paul G. Allen, founder of the Allen Institute for Brain Science, for his vision, encouragement and support.}}

\titlerunning{Automated neuron shape analysis from Electron Microscopy}

\author{S. Seshamani\inst{1} \and L. Elabbady\inst{1} \and C. Schneider-Mizell\inst{1} \and G. Mahalingam\inst{1} \and S. Dorkenwald\inst{2} \and A. L. Bodor\inst{1} \and T. Macrina\inst{2} \and D. J. Bumbarger\inst{1} \and J. Buchanan\inst{1} \and M. M. Takeno\inst{1} \and W. Yin\inst{1}, D. Brittain\inst{1}, R. Torres\inst{1} \and D. Kapner\inst{1} \and K. Lee\inst{2} \and R. Lu\inst{2} \and J. Wu\inst{2} \and N. M. Da Costa\inst{1} \and C.R. Reid\inst{1} \and F. Collman\inst{1}}

\institute{Allen Institute of Brain Science, Seattle WA, 
USA  \and
Princeton University, Princeton NJ 08544, USA }

\maketitle              

\begin{abstract}
Morphology based analysis of cell types has been an area of great interest to the neuroscience community for several decades. Recently, high resolution electron microscopy (EM) datasets of the mouse brain have opened up opportunities for data analysis at a level of detail that was previously impossible. These datasets are very large in nature and thus, manual analysis is not a practical solution. Of particular interest are details to the level of post synaptic structures. This paper proposes a fully automated framework for analysis of post-synaptic structure based neuron analysis from EM data. The processing framework involves shape extraction, representation with an autoencoder, and whole cell modeling and analysis based on shape distributions. We apply our novel framework on a dataset of 1031 neurons obtained from imaging a 1mm x 1mm x 40 micrometer volume of the mouse visual cortex and show the strength of our method in clustering and classification of neuronal shapes.

\end{abstract}

\section{Introduction}
\label{sec:intro}
Neurons receive thousands of synaptic inputs all over their cell body, axon initial segment, and dendrites,
and a \textbf{Postsynaptic Shape} (PSS) for a given synapse is the local morphological structure near its contact point. (Figure 1).
\begin{figure}[t]
    \centering
    \includegraphics[width=7cm]{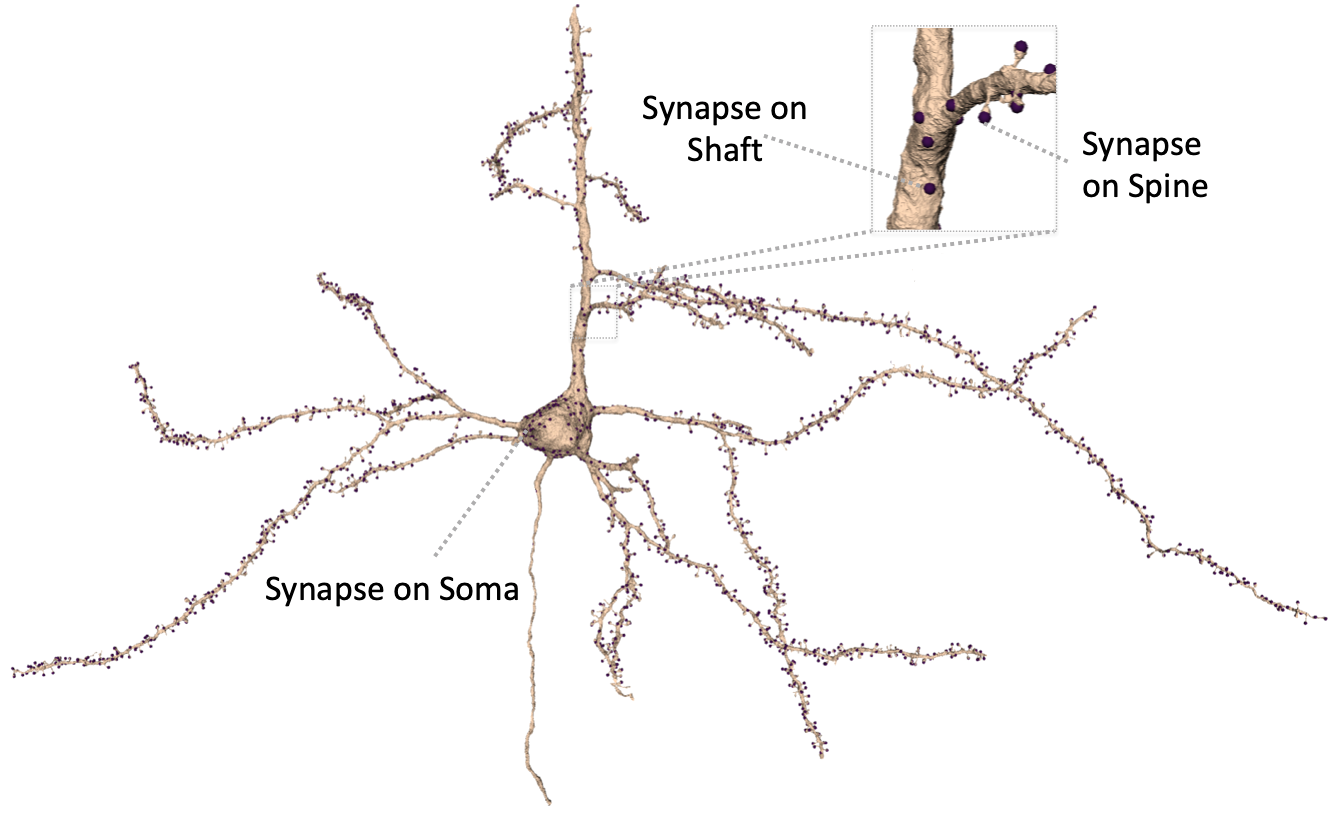}
    \includegraphics[width=3cm]{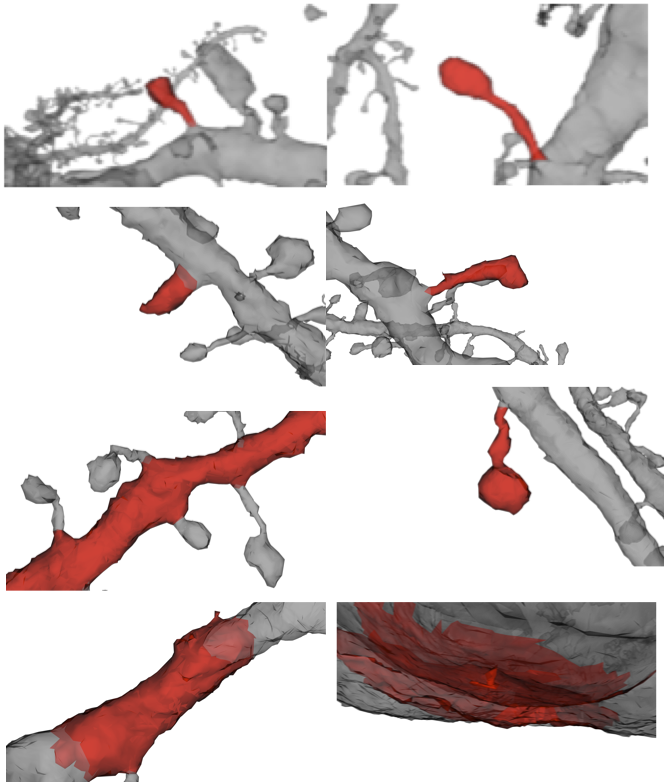}
    \caption{Left: Excitatory neuron (cream), with synapse locations (purple), examples of soma(cell body), spine and shaft synapses. Right: Examples of different PSS which include spines, shaft regions and soma regions}
    \label{fig:introimage}
\end{figure}
Excitatory neurons in cortex are described as being 'spiny' as most of the excitatory connections onto those cells occur on protrusions called spines. Inhibitory neurons are described as being 'aspiny', as most inputs onto these cells go directly onto dendritic shafts.  However, there are many sub-types of neurons beyond those simple distinctions \cite{Tasic2018,Gouwens2019}, and a quantitative assessment of the fine structures could be an effective way to lead to the discovery of these cell types. 
Recent advances in electron microscopy (EM) imaging technology and computer assisted reconstruction have enabled acquisition and analysis of increasingly large numbers of highly detailed 3D reconstructions of neuronal circuitry. These data have tremendous structural detail, and reveal the precise locations of synaptic connectivity between many neurons, along with the detailed morphologies of cellular structure. Previous methods to quantify fine neuronal structures have taken a manual labelling and/or a qualitative categorical description \cite{Chirillo2019}. However, to properly leverage this data into insights, there is a need for techniques that can automatically distill and summarize information in a biologically meaningful fashion, as there is simply too much data to consider by hand.


Data driven analyses of 3D meshes rely on the ability to extract a shape representation that will enable distinguishing between cell morphologies. Sholl analysis \cite{Sholl2015} is a method that has been commonly used in the literature for modelling cell morphologies based on only branch points in neuron skeletons. When it comes to EM cell meshes, down-sampled or skeleton representations would not retain fine morphological information and entire mesh representations would be too large, making comparisons computationally expensive. There is therefore a great need to develop a compact cell descriptor that retains fine morphology information. 
\cite{Schubert2019} proposed a compact neuron representation generated by rendering 3D rotations of neuronal components, processing them as images and then mapping the results back to the mesh. This however suffers from potential information loss due to multiple re-mappings.
In this paper, we propose a novel compact representation for mesh-based data-driven fine morphological clustering.
The method depends on automated synapse detection procedures \cite{synapsedetection} which enable PSS extraction from the local neighborhood. The main contributions of the paper are the following:
\begin{enumerate}
\item A method for describing and extracting local shape patterns (PSS) in a neuronal mesh
\item A new morphological descriptor of whole neurons
\item An automated framework for morphological analysis of whole neurons that can work on incomplete data using fine morphological features, and 
\item A validation of this framework on a large EM dataset.
\end{enumerate}
\section{Methods}
The objective is to generate a compact representation that enables comparison of fine morphological structures (PSS) between two neurons. PSS identification/extraction requires mesh localization for identifying the subset of vertices that form the PSS. We propose a synapse-based technique (Section 2.1) for performing this. For generating a descriptor for a whole cell, we further propose a framework consisting of two subcomponents: model generation and cell descriptor generation outlined in section 2.2 and 2.3. 
\begin{figure}[t]
    \centering
    \includegraphics[width=10cm]{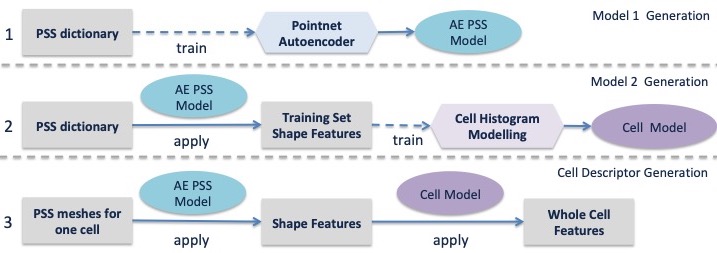}
    \caption{The framework comprises of : \textbf{Model Generation} and \textbf{Cell Descriptor Generation}.
Model Generation: we use a training set consisting of $N$ cells and extract all PSS from the cells to form a dictionary.
This is used to train a pointnet autoencoder which generates the AE PSS model.
The AE PSS model is applied on all training shapes to generate a set of features, which are used to learn a Cell Histogram Model with a codebook.
Cell Descriptor Generation: for each new cell, all postsynaptic shapes are first extracted.
Next, the trained AE PSS Model is applied to obtain shape features, and finally the cell histogram model is applied to obtain a cell descriptor. }
    \label{fig:workflowimage}
    
\end{figure}

\noindent \subsection{\textbf{Post-Synaptic Shape Extraction}}
With the advancement of automated synapse detection procedures \cite{synapsedetection}, it has been possible to extract locations of synapses within an EM volume with high accuracy. This enables synapse aware techniques for extracting global cell information like skeletons \cite{matejek2019synapseaware}. Here however, we use synapses for extracting local information around them. 
Given a cell mesh $M$ and a set of locations of synapses onto this cell $S_M = \{ s_1,s_2 ...\}$,  our algorithm first calculates a skeleton of the cell running along the shaft of the cell. Then for each synapse, it obtains a local region mesh falling within a fixed radius and segments it based on its Shape Diameter Function (SDF) into multiple sub-meshes. We then find the point on the local region mesh that is closest to the synapse and calculate the shortest path between this point and the skeleton. Finally, we find the sub-mesh that lies closest to the synapse and merge it will all other sub-meshes that fall on the shortest path and stop once the average SDF of the segment is greater than that of the previous one indicating that the spine has been traversed and the shaft has been reached. 


\noindent\textbf{Mesh Skeletonization:} This approach, related to TEASAR \cite{883951} and its modifications, is performed on the mesh graph, rather than on the voxel graph, for speed and memory use. The algorithm takes in a triangulated neuron mesh and treats it as an undirected graph, with vertices as nodes and mesh edges weighted by distance. It first calculates connected components to get a set of Vertex-Edge pairs as: $C = \{(V_i,E_i),... \}$.  The root node $v_{root}$ is calculated by finding the maximally far point along the mesh graph by bouncing from farthest point to farthest point and is used to initialize the skeleton. The rest of the skeleton is computed by repeatedly finding the farthest point to the skeleton, adding the shortest path from the current skeleton to this point and eliminating all other points within a distance threshold $d$ which is empirically selected as $d = 12000 nm$. This results in the skeleton running mostly along the shaft of the dendrites without reaching into the finer structures such as spines.  






\noindent \textbf{Surface Mesh Segmentation:}
\label{sec:segmentation}
We use a Shape Diameter Function (SDF) based algorithm  carried out with the CGAL surface segmentation package 
with the following parameters - number of clusters: 5, cone angle: $\pi/4$, number of rays: 5, smoothness: 0.3. This technique is quite sensitive to the size and detail of the surface mesh  and hence, we always sample a local mesh around a synapse (Algorithm 1, Line 6) with a fixed radius size of 3500 nm (picked empirically to ensure that large spines and at least some parts of the dendritic shaft and skeleton would be present). 




\subsection{\textbf{Model Generation }}
Model generation is carried out in 2 steps (Fig. 2). In the first, we present a method for generating local structure (PSS) encodings  and in the second, we propose a method that uses the encodings for describing whole cells.

\noindent \textbf{Local Structure Representations} 
To facilitate comparison across shapes and development of shape models, we develop a fixed size encoding for representing each 3D mesh of a PSS.  One way to obtain this is by extracting shape features to populate a descriptor. The choice of features however greatly impacts the information that the representation encodes. Recently, the pointnet architecture \cite{qi2017pointnet}  has been proposed for 3D point based autoencoders which circumvents user dependent feature selection. This architecture is made up of an encoder which reduces the dimensionality of the input 3D points into a fixed size vector and a decoder that takes this vector and outputs a set of 3D points. The learning step optimizes the differences between the input points and the output points and the learned model can generate a compact, fixed size encoding of a 3D point set.

Using deep learning for feature extraction however requires a training set. For this, we use a selected set of neurons and extract all its PSS to form a PSS shape dictionary.
Using shapes in this dictionary, our framework uses the pointnet autoencoder \cite{qi2017pointnet} to learn the desired fixed size representation. 
Input meshes are resampled with a fixed number of points, as required by the training procedure. Since our shapes are small, the pointnet (instead of pointnet++) architecture was sufficient to capture the complexities of our shapes.

\noindent \textbf{Whole Cell Histogram Model}
Given a set of mesh-feature pairs for all PSS in the dictionary, we propose to use a combination of Bag of Words (BoW) and Sholl techniques to describe cells. Bag of Words (BoW) techniques for feature space binning have been commonly used for image classification \cite{bow2005} and Sholl analysis is a classical method based on counting the number of branch points binned by distances from the cell body \cite{Sholl2015}. For the proposed model, we first use the fixed size encodings of all PSS in the training set, to estimate $k$ centers using k-means clustering. This gives the $k$ defining codewords in the BoW model. Additionally, we create $n$ bins for each of the $k$ codewords which represent $n$ radiuses from the center of the cell body for a total of $Q = kn$ bins for the cell codebook.

\subsection{\textbf{Cell Descriptor Generation }}
We use the models learned in the previous step to generate a cell descriptor (Fig. 2). 
For a new cell mesh, we extract all its PSS and apply the pointnet autoencoder model to obtain shape features. Using the 3D meshes of the PSS and its shape features, we bin the shapes into the $Q$ bins of the cell histogram model with the following steps. For each PSS described by a (mesh,feature) pair:
\begin{enumerate}
    \item Compute the distance of the mesh centroid and the center of the cell body. 
    \item Assign the PSS to the $n$th radius bin
    \item Compute Euclidean distances between each feature vector and BoW codewords 
    \item Assign each feature vector to the codeword that it is closest to. Combining with the $nth$ radius bin, this results in an assignment of each PSS into its $k-n$th bin.
\end{enumerate}
The descriptor proposed above can handle comparisons between cells at similar levels of completion. However, for EM data it is impossible to predict how complete a cell would be in the volume block that was imaged. The volume or surface area of the cell mesh can be used as an approximate measuring tool of completeness. In this work, we use the surface area measurement of the whole cell for normalizing the final cell histograms.

\section{Experiments}
\label{sec:pagestyle}
\noindent \textbf{Data} We analyzed an EM volume from mouse primary visual cortex which spanned all layers of the cortex and measured approximately 1x1x0.04 mm imaged at 4x4x40 nm/voxel resolution. Synapses were detected within the volume using an independent synapse detector using methods described in \cite{synapsedetection}. The volume includes 40 million synapses and 4235 cell bodies, of which 1031 had complete and accurate cell body (spherical part of the neuron that contains the nucleus) reconstructions which we use for our analysis. Cell body locations were manually labelled. Note that although the cell body was fully reconstructed, reconstruction of the whole cell was not obtained in any cases since the volume was cut off.
%
To ensure that we had a morphologically diverse training set that covered all cortical layers, categorical labels were given based on cortical depth and assigned by each cell body's distance from the cortical surface (Categories: Layer 2/3, Layer 4, Layer 5, Layer 6, Layer 2/3-4 Inhibitory, Layer 5/6 Inhibitory). 40 cells were randomly selected from each category, with one category (Lower Inhibitory) that had only 36 cells, for a total of 236 cells. 
For all these cells, we extracted all PSS (160,000 shapes) to form the PSS Shape Dictionary.

\begin{figure}[t]
    \centering
    \includegraphics[width=7
    cm]{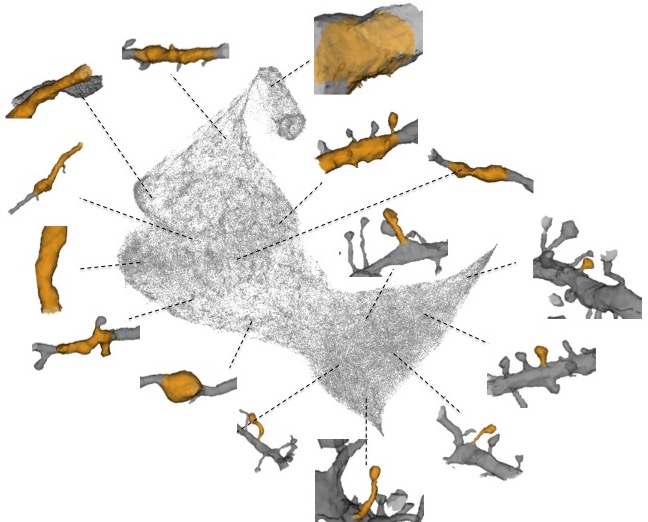}
    \caption{UMAP of 160k PSS shape features from 236 cells in the training set. Observe that the feature space distinguishes spines, cell body and shaft shapes.}
    \label{fig:umapspines}
\end{figure}
\noindent \textbf{Model Generation}
From training cells, we obtained 160k PSS, and computed encodings of dimension 1024 for all shapes with a pointnet autoencoder. We then applied UMAP \cite{2018arXivUMAP} to the data to visualize it in 2 dimensions. Fig. \ref{fig:umapspines} shows the range of the PSS space and how the local morphology varies across it. Note that spines, shafts and soma appear distinctly in this space.
We then computed a cell histogram model using $k = 20$ and $n = 24$ to give us a feature of size 480 which we normalized by the cell surface mesh area. 

\noindent \textbf{Evaluation}
For all cells, 2 independent labelers assigned categories to all 1031 cells manually: Inhibitory, Excitatory Layer 2/3, Excitatory Layer 4, Excitatory Layer 5 PT, Excitatory Layer 5 IT, Excitatory Layer 5 NP and Excitatory Layer 6. Additionally, labelers provided their confidence level on their rating: 0 for "Not Confident" and 1 for "Confident". After eliminating cells for which at least one labeler was not confident, we obtained a set of 532 cells. 
Finally, after eliminating cells where the labelers disagreed, we obtained a total of 505 cells that we used for our analysis.
We compared the proposed feature descriptor with two other feature descriptors and demonstrate the strength of the proposed framework in clustering and classification. Although there has been some extensive work such as \cite{Gouwens2019} which uses features extracted from skeletons for performing morphological analysis, most of these methods  depend on some manual annotation and cannot deal with incomplete cells. The most popularly used method for automated morphological analysis is Sholl Analysis \cite{Sholl2015}. The 3 cell description approaches we compare are:
\begin{enumerate}
    \item \textbf{Traditional Sholl Analysis with Branch Points (SBP)}: Baseline method, commonly used for automatic neuron morphology analysis. The number of branch points in the neuron mesh skeleton is counted at various radii from the center of the cell body. We use 24 radii ranging from 10000nm to 130000nm in steps of 5000nm.
    \item \textbf{Sholl Analysis with PSS Counts (SPC)}: The number of PSS as a feature and bin the features with the same 24 radii. This accounts for the density of synapses as a function of radius. (Steps 1 and 2 from Section 2.3)
    \item \textbf{Sholl Analysis with PSS Shape features (SPSF)}: Proposed feature with 480 sized cell model (Steps 1,2,3 and 4 from Section 2.3)
\end{enumerate}

The three methods we use for comparison include: K-means clustering, Phenograph clustering \cite{Levine2015Phenograph} (a k-nearest neighbors graph clustering algorithm) and SVM Classification. For K-means, we used $k=7$, the number of labels we obtained from the labelers. 
For Phenograph clustering, we used $k=23$. For the SVM, we used the rbf kernel, a polynomial degree of 3, the regularization parameter C = 1 and 10-fold cross validation. Table \ref{tab:withnorm} shows the rand index and mutual information values calculated between the predictions from each of these algorithms and the ground truth. We observe that the SPC outperforms SBP and SPSF further improves the performance with the shape information incorporated.

Different cortical cell types are found in different cortical layers. We thus investigated the effect of cell body location information on clustering and classification. For this, we used the depth of the cell from the surface of the cortex as an additional feature. This is a particularly useful feature since some of the sub-divisions in the labelling are based in part on this information. 
In order to incorporate the scalar value with comparable impact on each feature, we weighted it such that the distance metric $D$ between two cell histogram features $f_1$ and $f_2$ and depths $y_1$ and $y_2$ is calculated as:
$D(f_1,f_2,y_1,y_2) = dist(f_1,f_2) + \lambda |y_1-y_2|$
with $\lambda = 1/24 = 0.04166$, which was estimated empirically. Table \ref{tab:withnormandy} shows the clustering and classification results with the depth values incorporated. We observe a significant improvement in the performance of all algorithms with this information added and that the proposed method still consistently outperforms the other two.
\begin{table}
    \centering
    \begin{tabular}{|c|c|c|c|c|c|c|}
        \hline
        \textbf{Method} & \multicolumn{2}{|c|}{K-Means Clustering} & \multicolumn{2}{|c|}{Phenograph Clustering} & \multicolumn{2}{|c|}{SVM Classifier}  \\
        \hline
         & \textbf{RI} & \textbf{Mut Info} & \textbf{RI} & \textbf{Mut Info} & \textbf{RI} & \textbf{Mut Info}\\
        \hline
        SBP & 0.036 & 0.067 & 0.026 & 0.07 & 0.091 & 0.063 \\
        SPC & 0.188 & 0.295 & 0.199 & 0.306 & 0.285 & 0.32\\
        SPSF (Proposed)  & \textbf{0.387} & \textbf{0.435} & \textbf{0.388} & \textbf{0.483} & \textbf{0.709} & \textbf{0.654}\\
        \hline
    \end{tabular}
    \caption{Comparing the proposed descriptor to baseline methods with clustering and classification without depth information added. }
    \label{tab:withnorm}
\end{table}
\begin{figure}[ht]
    \centering
    \includegraphics[width=5.5cm]{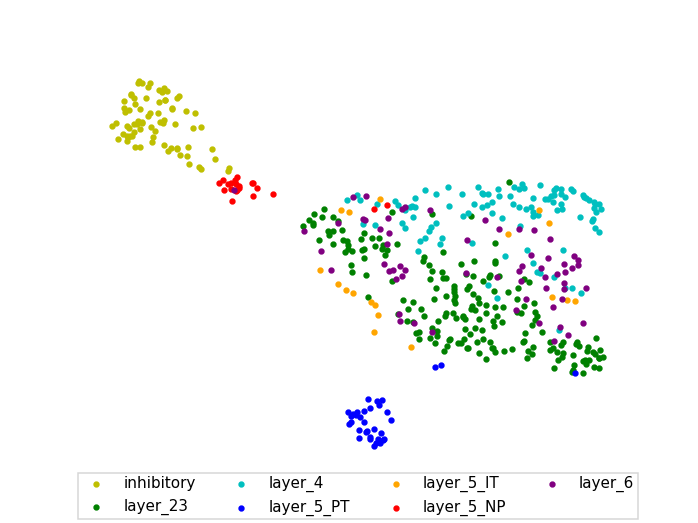}
    \includegraphics[width=5.5cm]{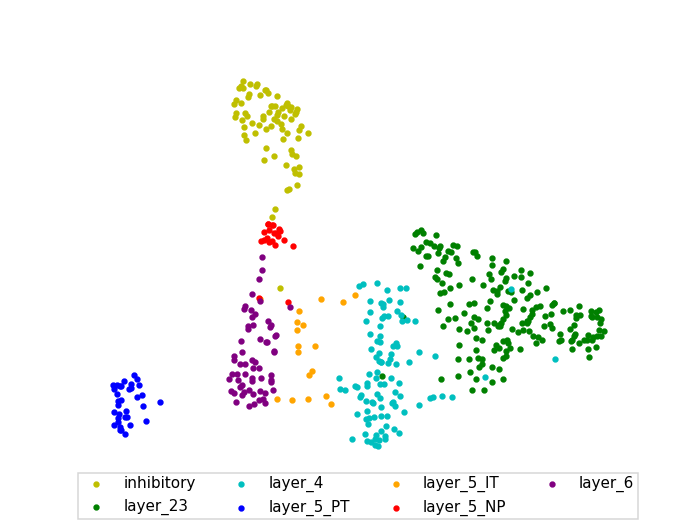}
    \caption{Applying UMAP to SPSF features for 505 cells without (left)  and With (rights) depth information. The feature distinguishes, inhibitory, layer 5 PT and layer 5 NP in both cases. Layer 2/3 and Layer 4 cells also separate well. However, Layer 5 IT and Layer 6 cells do overlap with these two classes when depth information is not used. This gets resolved when depth information is used.}
    \label{fig:resclustering}
\end{figure}

\begin{table}
    \centering
    \begin{tabular}{|c|c|c|c|c|c|c|}
        \hline
        \textbf{Method} & \multicolumn{2}{|c|}{K-Means Clustering} & \multicolumn{2}{|c|}{Phenograph Clustering} & \multicolumn{2}{|c|}{SVM Classifier}  \\
        \hline
         & \textbf{RI} & \textbf{Mut Info} & \textbf{RI} & \textbf{Mut Info} & \textbf{RI} & \textbf{Mut Info}\\
        \hline
        SBP & 0.133 & 0.181 & 0.131 & 0.204 & 0.632 & 0.472\\
        SPC & 0.425 & 0.495 &  0.497 & 0.607 & 0.833 & 0.752\\
        SPSF (Proposed) & \textbf{0.496} & \textbf{0.522} & \textbf{0.508} & \textbf{0.615} & \textbf{0.867} & \textbf{0.797}\\
        \hline
    \end{tabular}
    \caption{Comparing the proposed descriptor to baseline methods with clustering and classification with cell body depth information added. }
    \label{tab:withnormandy}
\end{table}

\section{Discussion and Conclusion}
\label{sec:typestyle}
We have presented a framework that enables automated fine morphological cell type clustering by leveraging the unique detail and scale of EM datasets. We validated our method by showing that we can distinguish between 7 classes of morphologically differing neurons and demonstrated the ability to perform unsupervised clustering and classification into groups based upon the diversity of PSS found on different types of neurons. What is especially remarkable is that our method could also handle cells with varying levels of completeness which even most semi-automatic neuron comparison methods have not been able to easily analyze. Current experiments have been performed on a set of 505 cells, but this technique can be immediately applied to larger and more complete datasets, making it a powerful tool for the description and discovery of neuronal sub-types.
The proposed features are currently being combined with other morphological information for further sub-type clustering and uncertainty based modeling for noisy and incomplete data.
This work has several potential applications in shape analysis of cell structure from superresolution microscopy and medical imaging of very large datasets where mesh representations provide compact, detailed shape information. Variations of this technique might also have uses in modalities where only 3D point clouds are generated, such as LiDAR.
\bibliographystyle{splncs04}
\bibliography{references}

\end{document}